\documentclass[journal]{IEEEtran}
\pdfoutput=1

\usepackage[utf8]{inputenc} 
\usepackage[T1]{fontenc}    
\usepackage{url}            
\usepackage{booktabs}       
\usepackage{amsfonts}       
\usepackage{nicefrac}       
\usepackage{microtype}      

\usepackage{mystyle}
\usepackage{subfigure}
\usepackage{graphicx}
\graphicspath{{images/}}

\usepackage{tikz}
\usepackage{pgfplots}
\pgfplotsset{compat=newest}
\usepackage{xintexpr}
\usepackage{lipsum}
\usepackage[ruled]{algorithm}
\usepackage{balance}

\usepackage[export]{adjustbox}

\newcommand{\stepref}[1]{{\it \ref{#1}}}

\allowdisplaybreaks

\begin{document}
\setlength{\textfloatsep}{1em}

\title{Accelerating GMM-based patch priors for image restoration: Three ingredients for a 100$\boldsymbol\times$ speed-up}

\author{Shibin Parameswaran,
  Charles-Alban Deledalle,
  Lo\"ic Denis
  and Truong~Q.~Nguyen
\thanks{S. Parameswaran, C.-A. Deledalle and T. Nguyen are with the Department
of Electrical and Computer Engineering, University of California, San Diego,
9500 Gilman Dr, La Jolla, CA 92093,
e-mail: sparames@ucsd.edu, cdeledalle@eng.ucsd.edu and tqn001@eng.ucsd.edu.}
\thanks{C. Deledalle is also with  IMB, CNRS, Univ. Bordeaux, Bordeaux INP,
  F-33405 Talence, France, e-mail: charles-alban.deledalle@math.u-bordeaux.fr}
\thanks{L. Denis is with Univ Lyon, UJM-Saint-Etienne, CNRS, Institut d Optique Graduate School, Laboratoire Hubert Curien UMR 5516, F-42023, Saint-Etienne, France,
  e-mail: loic.denis@univ-st-etienne.fr.}
}

\maketitle

\begin{abstract}
Image restoration methods aim to recover the underlying clean image
from corrupted observations. The Expected Patch Log-likelihood (EPLL)
algorithm is a powerful image restoration method that uses a Gaussian mixture model (GMM)
prior on the patches of natural images. Although it is very
effective for restoring images, its high runtime complexity makes EPLL ill-suited for
most practical applications. In this paper, we propose three approximations to the
original EPLL algorithm. The resulting algorithm, which we call the fast-EPLL (FEPLL), attains
a dramatic speed-up of two orders of magnitude over EPLL while
incurring a negligible drop in the
restored image quality (less than 0.5 dB).
We demonstrate the efficacy and versatility of our algorithm
on a number of inverse problems such as denoising, deblurring,
super-resolution, inpainting and devignetting. To the best of our knowledge, FEPLL is the first algorithm that can competitively restore a $512\!\times\!512$ pixel image in under 0.5s for \textit{all} the
degradations mentioned above \textit{without} specialized code optimizations such as CPU parallelization or GPU implementation.
\end{abstract}


\IEEEpeerreviewmaketitle

\section{Introduction}
Patch-based methods form a very popular and successful class of image restoration techniques. These methods
process an image on a patch-by-patch basis where a patch is a small sub-image ({\it e.g.}, of $8\!\times\!8$ pixels) that captures both geometric and textural information.
Patch-based algorithms have been at the core of many state-of-the-art results obtained on various image restoration problems such as denoising, deblurring,
super-resolution, defogging, or compression artifact removal to name a few.  In image
denoising, patch-based processing became popular after the success of the Non-Local Means
algorithm \cite{Buades05}. Subsequently, continued research efforts have led to
significant algorithmic advancements in this area
\cite{Aharon2006, Dabov07,Zoran11,  deledalle2011image, mairal2012task, yu2012solving, lebrun2013nonlocal}.
Other inverse problems such as image super-resolution and image deblurring have also benefited from
patch-based models
\cite{danielyan2008image, Glasner09, protter2009generalizing, singh2014sr, egiazarian2015single, lopez2016sr}.

Among these various patch-based
methods, the Expected Patch Log-Likelihood algorithm (EPLL) \cite{Zoran11} deserves a special mention
due to its restoration performance and versatility. The EPLL introduced an innovative application of
Gaussian Mixture Models (GMMs) to capture the prior distribution of patches in natural images.
Note that a similar idea was introduced concurrently in \cite{yu2012solving}.
The success of this method is evident from the large number of recent works that extend the
original EPLL formulation \cite{sulam2015expected,luo2016adaptive,cai2016image,papyan2016multi,ren2016example,houdard2017}.
However, a persistent problem of EPLL-based algorithms is their high
runtime complexity. For instance, it is orders of magnitude slower than the
well-engineered BM3D image denoising algorithm \cite{Dabov07}. However, extensions of
BM3D that perform super-resolution \cite{danielyan2010spatially} and other inverse problems \cite{katkovnik2009nonlocal}
require fundamental algorithmic changes, making BM3D far less adaptable than EPLL. Other approaches that are as versatile as EPLL \cite{schmidt2014shrinkage,chen2014insights,Jancsary2012regression} either
lack the algorithmic efficiency of BM3D or the restoration efficacy of EPLL.

Another class of techniques that arguably offers better runtime performance than EPLL-based methods
(but not BM3D) are those based on deep learning. With the advancements in computational resources,
researchers have attempted to solve
some classical inverse problems using multi-layer perceptrons \cite{burger2012}
and deep networks \cite{chen2015learning, dong2016image, kim2016accurate}.
These methods achieve very good restoration performance, but are
heavily dependent on the amount of training data available for each degradation scenario.
Most of these methods learn filters that are suited to restore a specific
noise level (denoising), blur (deblurring) or upsampling factor (super-resolution), which
makes them less attractive to serve as generic image restoration solutions. More recently,
Zhang {\it et al.}~\cite{zhang2016beyond} demonstrated the use of deep residual networks
for general denoising problems, single-image super-resolution and compression artifact removal. Unlike earlier
deep learning efforts, their approach can restore images with different noise levels using a single model which is learned by
training on image patches containing a range of degradations.
Even in this case, the underlying deep learning model requires retraining whenever a new
degradation scenario different from those considered during the learning stage is encountered.
Moreover, it is much harder to gain insight into the actual model learned by a deep architecture compared to a GMM.
For this reason, even with the advent of deep learning methods, flexible algorithms like EPLL that have a transparent formulation remain relevant for image restoration.

Recently, researchers have tried to improve the speed of EPLL by replacing the most time-consuming operation
in the EPLL algorithm with a machine learning-based technique of their choice \cite{wang2014discriminative,rosenbaum2015return}.
These methods were successful in accelerating EPLL to an extent but did not consider tackling all of its
bottlenecks. In contrast, this paper focuses on accelerating EPLL by
proposing algorithmic approximations to all the prospective bottlenecks present in the
original algorithm proposed by Zoran {\it et al.}~\cite{Zoran11}. To this end, we first provide
a complete computational and runtime analysis of EPLL, present a new and efficient
implementation of original EPLL algorithm and then finally propose innovative approximations that
lead to a novel algorithm that is more than 100$\times$ faster compared to the efficiently implemented EPLL (and
350$\times$ faster than the runtime obtained by using the original implementation \cite{Zoran11}).

\paragraph*{Contributions} The main contributions of this work are the following. We introduce three
strategies to accelerate patch-based image restoration algorithms that use a GMM prior. We show that, when used jointly,
they lead to a speed-up of the EPLL algorithm
by two orders
of magnitude. Compared to the popular
BM3D algorithm, which represents the current state-of-the-art in terms of speed among CPU-based implementations,
the proposed algorithm is almost an order of magnitude faster. The three
strategies introduced in this work are general enough to be  applied individually or in any combination to accelerate
other related algorithms. For example, the random subsampling strategy is a general
technique that could be reused in any algorithm that considers overlapping patches to process images; the flat tail spectrum
approximation can accelerate any method that needs Gaussian log-likelihood or
multiple Mahalanobis metric calculations; finally, the
binary search tree for Gaussian matching can be included in any algorithm based on a GMM prior model
and can be easily adapted for vector quantization techniques that use a dictionary.

For reproducibility purposes, we release our software on GitHub along with a few usage demonstrations (available at \url{https://goo.gl/xjqKUA}).

\section{Expected Patch Log-Likelihood (EPLL)}
We consider the problem of estimating an image
$\bx \in \RR^N$ ($N$ is the number of pixels)
from noisy linear observations $\by = \Aa \bx + \bw$, where
$\Aa: \RR^N \to \RR^M$ is a linear operator
and $\bw \in \RR^M$ is a noise component assumed to be white and Gaussian with variance $\sigma^2$.
In a standard denoising problem $\Aa$ is the identity matrix, but in more general settings,
it can account for loss of information or blurring.
Typical examples for operator $\Aa$ are: a low pass filter (for {\it deconvolution}),
a masking operator (for {\it inpainting}), or a projection
on a random subspace
(for {\it compressive sensing}).
To reduce noise and stabilize the inversion of $\Aa$,
some \textit{prior} information is used for the estimation of $\bx$.
The EPLL introduced by Zoran and Weiss \cite{Zoran11} includes this {\it prior} information as a model for the distribution of patches found in natural images.
Specifically, the EPLL defines the restored image as the maximum {\it{a posteriori}} estimate, corresponding to
the following minimization problem:
\begin{align}\label{eq:epll}
  \uargmin{\bx} \frac{P}{2 \sigma^2} \norm{\Aa \bx - \by}^2 - \sum_{i\in\Ii} \log p\left( \Pp_i \bx \right)
\end{align}
where $\Ii = \{1, \ldots, N\}$ is the set of pixel indices, $\Pp_i : \RR^N \to \RR^{P}$ is the linear operator
extracting a patch with $P$ pixels centered at the pixel with location $i$ (typically, $P=8 \!\times\! 8$),
and $p(.)$ is the {\it a priori} probability density function (\textit{i.e.}, the statistical model of noiseless patches in natural images).
While the first term in eq.~\eqref{eq:epll} ensures that
$\Aa \bx$ is close to the observations $\by$
(this term is the negative log-likelihood under the white Gaussian noise assumption),
the second term regularizes the solution $\bx$ by favoring an image such that all its patches fit the {\it prior} model of patches in natural images.
The authors of \cite{Zoran11} showed that this {\it prior} can be well approximated (upon removal of the DC component of each patch) using a zero-mean
Gaussian Mixture Model (GMM) with $K\!=\!200$ components, that reads
for any patch $\bz \in \RR^{P}$, as
\begin{align}
  p\left( \bz \right) =
  \sum_{k=1}^K w_k \frac1{(2\pi)^{P/2} |\bSigma_k|^{1/2}}\exp\left(-\frac12 \bz^t \bSigma_k^{-1} \bz\right),
  \label{eq:prior}
\end{align}
where the weights $w_k$ (such that $w_k\!>\!0$ and $\sum_k w_k\!=\!1$)
and the covariance matrices $\bSigma_k \in \RR^{P \times P}$
are estimated using the Expectation-Maximization algorithm \cite{dempster1977maximum}
on a dataset consisting of 2 million ``clean'' patches
extracted from the training set of the Berkeley Segmentation (BSDS) dataset \cite{martin2001database}.

\newcommand{\barplot}[2]{
  \pgfmathprintnumber[fixed, fixed zerofill,precision=2]{\xintthefloatexpr #1 \relax}s
  &
  {\tikz\draw [fill=black!50] (0,0) rectangle ({(#1)/(#2) * 3.4},0.2);}
  \pgfmathprintnumber[fixed,precision=0]{\xintthefloatexpr 100*(#1)/(#2) \relax} \%
}
\begin{table*}
  \caption{Comparison of the execution time of our implementation of EPLL with
  and without proposed accelerations.
    Experiment conducted on a $481 \times 321$ image denoising problem.
    Profiling was carried out using {\texttt{MATLAB}} (R2014b) on
    a PC with Intel(R) Core(TM) i7-4790K CPU @4.00GHz and 16 GB RAM.
    Execution times are reported in seconds (s) and as a percentage of the total time (\%).
  }
  \label{tab:profiling}
  \centering
  \begin{tabular}{@{}lrlrl@{}}
    \hline
    Step & \multicolumn{2}{l}{Without accelerations} & \multicolumn{2}{l}{With the proposed accelerations}\\
    \hline
    \hline
    \eqref{eq:gauss_select}
    &
    \barplot{43.614-0.08}{45.694}
    &
    \barplot{0.2316}{0.3499}
    \\
    \eqref{eq:patchest}
    &
    \barplot{0.453+0.373+0.128}{45.694}
    &
    \barplot{0.0082+0.0088+0.0093+0.0200}{0.3499}
    \\
    \eqref{eq:patch_extraction}
    &
    \barplot{0.258+0.198}{45.694}
    &
    \barplot{0.0146+0.0104}{0.3499}
    \\
    \eqref{eq:patch_reprojection}
    &
    \barplot{0.153+0.08}{45.694}
    &
    \barplot{0.0123}{0.3499}
    \\
    Others
    &
    \barplot{45.694-45.177}{45.694}
    &
    \barplot{(1-0.6619-0.1323-0.0714-0.035)*0.3499}{0.3499}\\
    \hline
    Total
    &
    \hfill 45.69s
    &
    &
    \hfill 0.35s
    &
    \\
    \hline
  \end{tabular}
  \vskip1em
\end{table*}

\paragraph*{Half-quadratic splitting}
Problem \eqref{eq:epll} is a large non-convex problem where $\Aa$ couples all unknown pixel values $\bx$ and the patch prior is highly non-convex.
A classical workaround, known as half-quadratic splitting \cite{geman1995nonlinear,krishnan2009fast},
is to introduce $N$ auxiliary unknown vectors $\bz_i \in \RR^P$, and consider instead
the penalized optimization problem that reads, for $\beta > 0$, as
\begin{multline}\label{eq:qhs}
  \uargmin{\bx, \bz_1, \ldots, \bz_N}
  \frac{P}{2 \sigma^2} \norm{\Aa \bx - \by}^2\\
  + \frac{\beta}{2} \sum_{i\in \Ii} \norm{\Pp_i \bx - \bz_i}^2
  -  \sum_{i\in \Ii}\log p\left( \bz_i \right).
\end{multline}
When $\beta \to \infty$, the problem \eqref{eq:qhs} becomes equivalent to the original problem \eqref{eq:epll}. In practice, an increasing sequence of $\beta$ is considered, and an alternating optimization scheme is used:
\begin{align}
    \label{eq:qhs_optim_paest}
  \biggl\{\hat \bz_i &\leftarrow \uargmin{\bz_i}
  \frac{\beta}{2} \norm{\Pp_i \hat \bx - \bz_i}^2 - \log p\left( \bz_i \right)\biggr\}_{i=1..N}\\
  \label{eq:qhs_optim_imest}
  \hat \bx &\leftarrow \uargmin{\bx}
  \frac{P}{2 \sigma^2} \norm{\Aa \bx - \by}^2
  + \frac{\beta}{2} \sum_{i\in \Ii} \norm{\Pp_i \bx - \hat \bz_i}^2.
\end{align}
\begin{algorithm}[t]
  \small
      {\hspace{-3em}\begin{minipage}{1.11\linewidth}\vspace{-1em}\begin{align}
    &\text{for all } i \in \Ii \nonumber
    \\
    \label{eq:patch_extraction}
    \tag{Patch extraction}
    &\makebox(0,0){\hspace*{3.5ex}\rule[-32ex]{.2mm}{18ex}}\quad\quad
    \tilde \bz_i \leftarrow \Pp_i \hat\bx
    \\
    &\quad\quad
    k_i^\star
    \leftarrow \uargmin{1 \leq k_i \leq K}
    \log w_{k_i}^{-2} + \log \left|\bSigma_{k_i} \!+\! \tfrac{1}{\beta} \Id_P\right| +
    \nonumber
    \\
    \label{eq:gauss_select}
    \tag{Gaussian selection}
    &
    \quad\quad\quad\quad\quad\quad\quad\quad\quad
    \tilde \bz_i^t \left(\bSigma_{k_i} \!+\! \tfrac{1}{\beta} \Id_P \right)^{-1} \!\!\tilde \bz_i
    \hspace{-10em}
    \\
    \label{eq:patchest}
    \tag{Patch estimation}
    &\quad\quad
    \hat{\bz}_i
    \leftarrow
    \left(\bSigma_{k_i^\star} + \tfrac{1}{\beta} \Id_P\right)^{-1} \bSigma_{k_i^\star} \tilde \bz_i
    \\
    \label{eq:patch_reprojection}
    \tag{Patch reprojection}
    &\tilde \bx \leftarrow
    \big(\text{\small ${\sum_{i\in \Ii}}$} \Pp_i^t \Pp_i\big)^{-1}
    \text{\small ${\sum_{i\in \Ii}}$} \Pp_i^t \hat \bz_i
    \\
    \label{eq:imest}
    \tag{Image estimation}
    &\hat \bx \leftarrow
    \big(
    \Aa^t \Aa
    + \beta \sigma^2
    \Id_N
    \big)^{-1}
    \left(
    \Aa^t \by
    +
    \beta \sigma^2 \tilde \bx
    \right)
    \end{align}\end{minipage}%
  }%
  \caption{The five steps of an EPLL iteration \cite{Zoran11}}
  \label{alg:epll}
\end{algorithm}%
\paragraph*{Algorithm}
Subproblem \eqref{eq:qhs_optim_imest} corresponds to solving a linear inverse problem with a Tikhonov regularization,
and has an explicit solution often referred to as Wiener filtering:
\begin{align}\label{eq:qhs_optim_imest_sol}
  \hat{\bx} = \left(\Aa^t\Aa+\frac{\beta\sigma^2}{P}\sum_{i \in \Ii}\Pp_i^t\Pp_i\right)^{-1} \left(\mathcal{A}^t \by + \frac{\beta\sigma^2}{P}\sum_{i \in \Ii}\Pp_i^t \hat \bz_i\right),
\end{align}
where $\Pp_i^t\Pp_i$ is a diagonal matrix whose $i$-th diagonal element corresponds to
the number of patches overlapping the pixel of index $i$.
This number is a constant equal to $P$
(assuming proper boundary conditions are used),
which allows to split the computation into two steps
\stepref{eq:patch_reprojection} and \stepref{eq:imest} as shown in Alg.~\ref{alg:epll}.
Note that the step \stepref{eq:patch_reprojection} is simply the average of all overlapping patches.
In contrast,
subproblem \eqref{eq:qhs_optim_paest} cannot be obtained in closed form
as it involves a term with the logarithm of a sum of exponentials.
A practical solution proposed in \cite{Zoran11} is to keep only
the component $k_i^\star$ maximizing
the likelihood for the given patch assuming it is
a zero-mean Gaussian random vector with covariance matrix
$\bSigma_{k_i} + \tfrac{1}{\beta} \Id_P$.
With this approximation, the
solution of \eqref{eq:qhs_optim_paest} is also given by Wiener filtering,
and the resulting algorithm iterates the steps described in Alg.~\ref{alg:epll}.
The authors of \cite{Zoran11} found that using $T\!=\!5$ iterations,
with the sequence $\beta = \frac{1}{\sigma^2} \{ 1, 4, 8, 16, 32 \}$, for the initialization
$\hat \bx = \by$,
provides relevant solutions in denoising contexts for a wide range of noise level values $\sigma^2$.
\subsection{Complexity via eigenspace implementation}
The algorithm summarized in the previous section may reveal cumbersome
computations as it requires performing numerous matrix multiplications and
inversions. Nevertheless, as the matrices $\bSigma_k$
are known {\it prior} to any calculation, their eigendecomposition can be
computed offline
to improve the runtime.
If we denote the eigendecomposition (obtained offline) of $\bSigma_k = \bU_k \bS_k \bU_k^t$,
such that $\bU_k \in \RR^{P \times P}$ is unitary and $\bS_k$ is diagonal with positive
diagonal elements ordered in decreasing order,
steps \stepref{eq:gauss_select} and \stepref{eq:patchest} can be expressed
in the space of coefficients $\bc$ as
\begin{align}
  \label{eq:projection_eigenspace}
  &\biggl\{\tilde \bc^k_i \leftarrow \bU_k^t \tilde \bz_i\biggr\}_{\substack{\\[-1em] k=1..K\\i=1..N}}
  && \Oo(N K P^2)
  \\
  &\hspace*{23.5ex}\makebox[0pt]{$\displaystyle\biggl\{k_i^\star \leftarrow \uargmin{1 \leq k \leq K}
  \iota^{k} + \sum_{j = 1}^P \biggl(\log \nu^{k}_j +
  \frac{[\tilde \bc^{k}_{i}]_j^2}{\nu^{k}_j}\biggr)\biggr\}_{i=1..N}$}\nonumber
  \hspace{6em}
  \\
\label{eq:selection_eigenspace}
&\hspace*{20ex}
  && \Oo(N K P)
  \\
  &\biggl\{[\hat{\bc}_i]_j
  \leftarrow
  \gamma^{k_i^\star}_j [\tilde \bc^{k_i^\star}_{i}]_j\biggr\}_
{\substack{\\[-1em] j=1..P\\i=1..N}}
    && \Oo(N P)
  \\
  &\biggl\{\hat \bz_i \leftarrow \bU_{k_i^\star} \hat{\bc}_i\biggr\}_{i=1..N}
  && \Oo(N P^2)
\end{align}
where $[\tilde{\bc}]_j$ denotes the $j$-th entry of vector $\tilde\bc$,
$\iota^{k} = -2 \log w_{k}$,
$\nu^{k}_{j} = [\bS_{k}]_{jj} \!+\! \tfrac{1}{\beta}$,
and $\gamma^{k}_j = [\bS_{k}]_{jj} / \nu^{k}_{j}$ with $[\bS_{k}]_{jj}$ the $j$-th entry on the diagonal of matrix $\bS_{k}$.
The complexity of each operation is indicated on its right and corresponds to
the number of operations per iteration of the alternate optimization scheme.
The steps \stepref{eq:patch_extraction} and \stepref{eq:patch_reprojection}
share a complexity of $\Oo(N P)$.
Finally,
the complexity of step \stepref{eq:imest} depends on the transform $\Aa$.
In many scenarios of interest, $\Aa^t \Aa$ can be diagonalized using a fast transform and the inversion of
$\Aa^t \Aa + \beta \sigma^2 \Id_N$ can be performed efficiently in
the transformed domain (since $\Id_N$ is diagonal in any orthonormal basis).
For instance, it leads to
$\Oo(N)$ operations for denoising or inpainting, and $\Oo(N\log N)$
for periodical deconvolutions or super-resolution problems,
thanks to the fast Fourier transform.
If $\Aa$ cannot be easily diagonalized,
 this step can be performed using conjugate gradient (CG) method, as done in \cite{Zoran11}, at a computational cost that depends on the number of CG iterations ({\it{i.e.}}, on the conditioning of $\Aa^t \Aa
    + \beta \sigma^2
    \Id_N$).
In any case, as shown in the next section, this step has a complexity independent of $P$ and $K$ and
is one of the faster operations in the image restoration problems considered in this paper.
\subsection{Computation time analysis}

In order to uncover the practical computational bottlenecks of EPLL,
we have performed the following computational analysis.
To identify clearly which part is time consuming,
it is important
to make the algorithm implementation as optimal as possible.
Therefore, we refrain from using the \texttt{MATLAB} code provided by the original authors \cite{Zoran11} for speed comparisons. Instead, we use a \texttt{MATLAB}/\texttt{C} version of EPLL based on the eigenspace implementation described above,
where some steps are written in \texttt{C} language
and interfaced using \texttt{mex} functions. This version, which we refer to as EPLLc, provides results identical
to the original implementation while being 2-3 times faster.
The execution time
of each step for a single run of EPLLc is reported in the second column of Table \ref{tab:profiling}.
Reported times fit our complexity analysis and clearly indicate that the
Step \stepref{eq:gauss_select} causes significant bottleneck due to $\Oo(N P^2 K)$ complexity.

\bigskip

In the next section, we propose three independent modifications
leading to an algorithm with a complexity of $\Oo( N P \bar{r} \log K / s^2 )$
with two constants $1 \leq s^2 \leq P$ and $1 \leq \bar{r} \leq P$ that control the accuracy of the approximations introduced. The algorithm,
in practice, is more than $100$ times faster as shown by
its runtimes reported in the third column.
\section{Fast EPLL: the three key ingredients}
We propose three accelerations based on (i)
scanning only a (random) subset of the $N$ patches, (ii)
reducing the number of mixture components matched,
and (iii) projecting on a smaller subspace of the covariance eigenspace.
We begin by describing this latter acceleration strategy in the following paragraph.
\begin{figure}[!t]
  \centering%
  \hfill\subfigure{\includegraphics[width=0.979\linewidth]{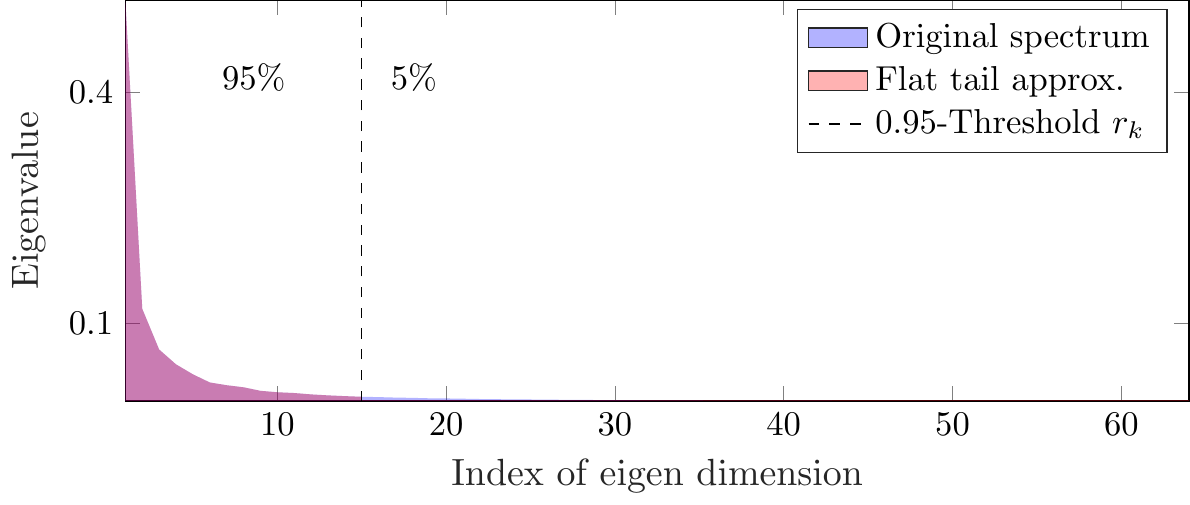}}\\
  \subfigure{\includegraphics[width=\linewidth]{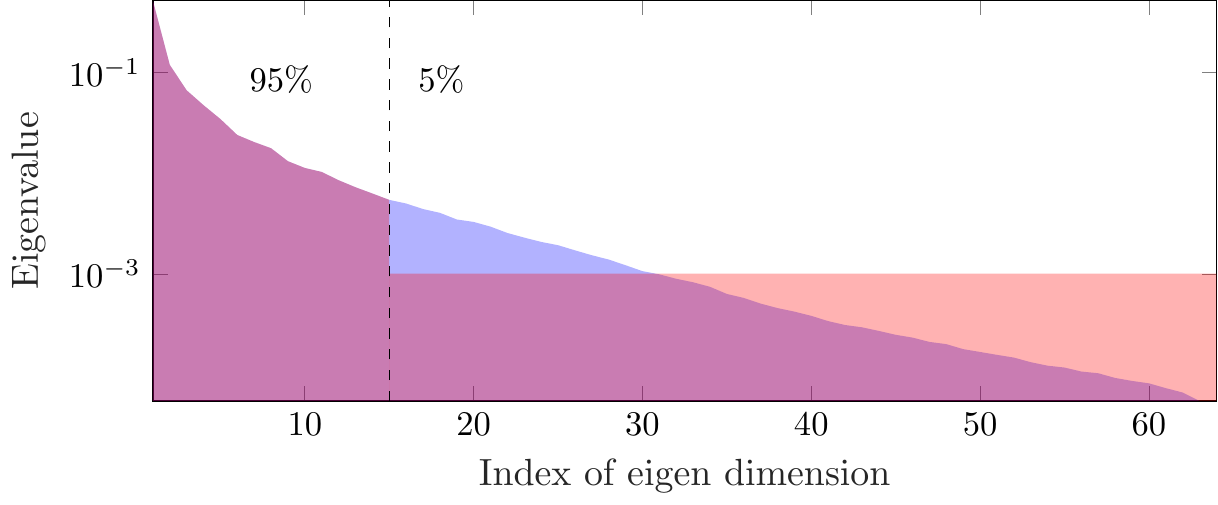}}\\
  \caption{Flat tail approximation: (a) with eigenvalues
    display on linear and (b) logarithmic scale.
    The $r_k = 15$ first eigen components explains
    $\rho\!=\!95\%$ of the variability.
  }
  \label{fig:flat_tail}
  \vskip1em
\end{figure}
\subsection{Speed-up via flat tail spectrum approximation}
To avoid computing the $P$ coefficients of the vector $\tilde c_i^k$ in
eq.~\eqref{eq:projection_eigenspace}, we rely on a flat-tail approximation.
The $k$-th Gaussian model is said to have a flat tail if there exists a rank $r_k$ such that for any $j\!>\!r_k$, the eigenvalues are constant: $[\bS_k]_{j,j}\!=\!\lambda_k$.
Denoting by $\bar\bU_k \!\in\! \RR^{P \times r_k}$ (resp.~$\bar\bU_k^c \!\in\! \RR^{P \times r_k^c}$)
the matrix formed by the $r_k$ first (resp. $r_k^c\!=\!P\!-\!r_k$ last) columns of $\bU_k$,
we have $\bar\bU_k^c (\bar\bU_k^c)^t \!=\! \Id_P \!-\! \bar\bU_k \bar\bU_k^t$.
It follows
\begin{align}
  (\bSigma_k + \tfrac1{\beta} \Id_P)^{-1}
  &=
  \bar\bU_k (\bar \bS_k + \tfrac1{\beta} \Id_{r_k})^{-1} \bar\bU_k^t
  \nonumber
  \\
  &
  + (\lambda_k + \tfrac1{\beta})^{-1} (\Id_P - \bar\bU_k \bar\bU_k^t),
  \\
  (\bSigma_k + \tfrac1{\beta} \Id_P)^{-1}
  \bSigma_k
  &=
  \bar\bU_k (\bar \bS_k + \tfrac1{\beta} \Id_{r_k})^{-1} \bar \bS_k \bar\bU_k^t
  \nonumber
  \\
  &
  + \lambda_k (\lambda_k + \tfrac1{\beta})^{-1} (
  \Id_P - \bar\bU_k \bar\bU_k^t),
\end{align}
where $\bar\bS_k \in \RR^{r_k \times r_k}$ is the diagonal matrix formed
by the $r_k$ first rows and columns of $\bS_k$.
Steps \stepref{eq:gauss_select} and \stepref{eq:patchest} can thus be rewritten as
\begin{align}
  &\biggl\{\tilde \bc^k_i \leftarrow \bar\bU_k^t \tilde \bz_i\biggr\}_{\substack{\\[-1em] k=1..K\\i=1..N}}
  && \!\!\!\!\!\Oo(N K P \bar r)
  \\
  &\biggl\{k_i^\star \leftarrow \uargmin{1 \leq k \leq K}
  \iota_{k} +
  r_k^c \log \nu^{k}_{P}
  \!+\!
  \frac{\norm{\tilde \bz_i}^2}{\nu^{k}_{P}}
  \nonumber\\
  &\hspace{0.3cm}
  \!+\!
  \sum_{j=1}^{r_k}\biggl(
  \log \nu^{k}_j
  \!+\!
  \frac{[\tilde \bc^{k}_{i}]_j^2}{\nu^{k}_j}
  \!-\!
  \frac{[\tilde \bc^{k}_{i}]_j^2}{\nu^{k}_{P}}\biggr)\biggr\}_{i=1..N}
  && \Oo(N K \bar r)
  \\
  & \biggl\{[\hat{\bc}_i]_j
  \leftarrow
  (\gamma^{k_i^\star}_{j} \!-\! \gamma^{k_i^\star}_{P}) \bigl[\tilde \bc^{k_i^\star}_{i}\bigr]_j\biggr\}_{\substack{\\[-1.2em] j=1..r_{k_i^\star}\\i=1..N\hspace*{.8ex}}}
  && \Oo(N P \bar r)
  \\
  &\biggl\{\hat \bz_i \leftarrow
  \bar \bU_{k_i^\star} \hat{\bc}_i
  \!+\!
  \gamma^{k_i^\star}_{P} \tilde \bz_i\biggr\}_{i=1..N}
  && \Oo(N P \bar r)
\end{align}
where
$\nu^{k}_{P} \!=\! \lambda_k \!+\! \tfrac{1}{\beta}$,
$\gamma^{k}_P \!=\! \lambda_k / \nu^{k}_{P}$.
As $\norm{\tilde \bz_i}^2$ can be computed once for all $k$,
the complexity of each step is
divided by $P/\bar r$, where $\bar r \!=\! \tfrac1K \sum_{k = 1}^K r_k$
is the average rank after which eigenvalues are considered constant.

In practice, covariance matrices $\bSigma_k$ are not flat-tail but can be approximated by a flat-tail matrix by replacing the lowest eigenvalues by a constant $\lambda_k$.
To obtain a small value of $\bar r$ (hence a large speed-up), we preserve a fixed proportion $\rho\!\in (0, 1]$ of the total variability and replace the smallest eigenvalues accounting for the remaining $1\!-\!\rho$ fraction of the variability by their average
(see Fig.~\ref{fig:flat_tail}): $r_k$ is the smallest integer such that $\Tr(\bar \bS_k)\geq \rho \Tr(\bS_k)$.
Choosing $\rho\!=\!0.95$ means that $5\%$ of the variability,
in the eigendirections associated to the smallest eigenvalues,
is assumed to be evenly spread in these directions.
In practice, the choice of $\rho\!=\!0.95$ leads to an average rank of $\bar r \!=\! 19.6$
(for $P\!=\!\,8\times 8\,$)
for a small drop of PSNR as shown in Fig.~\ref{fig:time_and_psnr}.
Among several other covariance approximations that we tested,
for instance, the one consisting in keeping only the $r_k$ first directions,
the flat tail approximation provided the best trade-off in terms
of acceleration and restoration quality.
\subsection{Speed-up via a balanced search tree}
\begin{figure}[!t]
  \centering%
  \includegraphics[width=0.99\linewidth,viewport=0 0 288 127,clip]{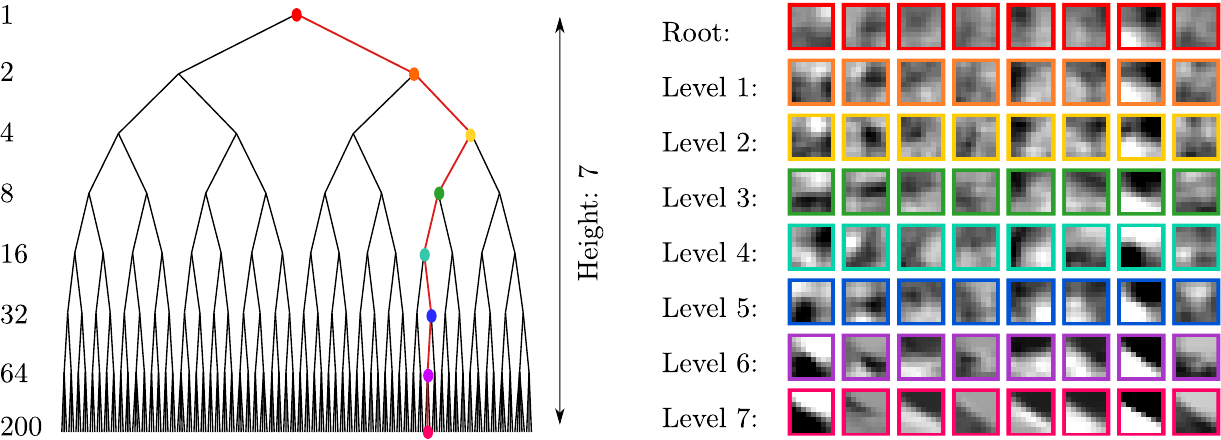}\\
  \caption{(left) Our obtained search tree (with numbers of nodes for each level).
    (right) Four patches well represented by each of the eight nodes along the branch highlighted in red in the tree. These patches are randomly sampled from the generative model
    encoded at each node of the branch (patches on the same column are generated from the same random seed).
  }
  \label{fig:bin_tree}
  \vskip1em
\end{figure}
As shown in Table \ref{tab:profiling},
the step \stepref{eq:gauss_select} has a complexity of $\Oo(N P^2 K)$, reduced to $\Oo(N P K \bar{r})$ using the flat tail spectrum approximation.
This step remains the biggest bottleneck
since each query patch has to be compared to all the
$K$ components of the GMM. To make this step even more
efficient, we reduce  its complexity using a balanced search tree. As described below,
such a tree can be built offline
by repeatedly collapsing the original GMM to models with fewer
components, until the entire model is reduced to a single Gaussian model.

We progressively combine the GMM components from one level to form the level above,
by clustering the $K$ components into $L<K$ clusters of similar ones,
until the entire model is reduced to a single component.
The similarity between two zero-mean Gaussian models with
covariance $\bSigma_1$ and $\bSigma_2$ is measured
by the symmetric Kullback-Leibler (KL) divergence
\begin{align}
\KL(\bSigma_1, \bSigma_2) = \tfrac{1}{2} \Tr(\bSigma_2^{-1}\bSigma_1 + \bSigma_1^{-1}\bSigma_2  - 2 \Id_P).
\end{align}
Based on this divergence, at each level $n$,
we look for a partition $\Omega^{\graphlevel{n}}$ of the $K$ Gaussian components
into $L$ clusters (with about equal sizes) minimizing the following
optimization problem
\begin{align}
  \label{eq:cost_clustering}
  \uargmin{\Omega^{\graphlevel{n}}}
  \sum_{l = 1}^L \sum_{{k_1, k_2 \in \Omega_l^{\graphlevel{n}}}} \KL(\bSigma_{k_1}, \bSigma_{k_1})
  ,
\end{align}
such that
$\bigcup_{l = 1}^L \Omega_l^{\graphlevel{n}} = [K]$
and
$\Omega_{l_1}^{\graphlevel{n}} \underset{\!l_1 \ne l_2\!}{\cap} \Omega_{l_2}^{\graphlevel{n}} = \emptyset$,
where $\Omega_l^{\graphlevel{n}}$ is the $l$-th set of Gaussian components
for the GMM at level $n$.
This clustering problem can be approximately solved
using the genetic algorithm of \cite{Joseph_Kirk_2014_omtsp} for the
Multiple Traveling Salesmen Problem (MTSP).
MTSP is a variation of the classical Traveling Salesman Problem where
several salesmen visit a unique set of cities and return to their origins,
and each city is visited by exactly one salesman. This attempts
to minimize the total distance traveled by all salesmen. Hence, it is similar
to our original problem given in eq.~\eqref{eq:cost_clustering} where the Gaussian
components and the clusters correspond to $K$ cities and $L$ salesmen, respectively.
Given the clustering at level $n$,
the new GMM at level $n-1$ is obtained by combining
the zero-mean Gaussian components such that, for all $1 \leq l \leq L$:
\begin{align}
w_{l}^{\graphlevel{n-1}} =  \sum\limits_{k\in\Omega_l^{\graphlevel{n}}} w_k^{\graphlevel{n}}
\qandq
\bSigma_{l}^{\graphlevel{n-1}} =  \frac{1
}{w_l^{\graphlevel{n-1}}} \sum\limits_{k\in\Omega_l^{\graphlevel{n}}}w_k^{\graphlevel{n}} \bSigma_k^{\graphlevel{n}},
\end{align}
where $\bSigma^{\graphlevel{n}}_k$ and  $w^{\graphlevel{n}}_k$ are the
corresponding covariance matrix and weight of the $k$-th Gaussian component
at level $n$.
Following this scheme,
the original GMM of $K\!=\!200$ components
is collapsed into increasingly more compact GMMs
with $K\!=\!64$, $32$, $16$, $8$, $4$, $2$ and $1$ components.
The main advantage of using MTSP compared to classical clustering approaches,
is that this procedure can be adapted easily to enforce approximately equal sized clusters,
simply by enforcing that each salesman visits at least
$3$ cities for the last level and $2$ for the other ones.

We also experimented with other clustering strategies such
as the hierarchical kmeans-like clustering in \cite{goldberger2004hierarchical} and
hierarchical agglomerative clustering. With no principled way to enforce even-sized
clusters, these approaches, in general, lead to
unbalanced trees (with comb structured branches) which result in large
variations in computation times from one image to another.
Although they all lead to similar denoising performances, we opted for
MTSP based clustering to build our Gaussian tree in favor of
obtaining a stable speed-up profile for
our resulting algorithm.

In Fig. \ref{fig:bin_tree} we show that the tree obtained using MTSP-based
clustering is almost a binary tree (left) and also display the types of patches
it encodes along a given path (right).
Such a balanced tree structure lets one avoid testing each patch against all
$K$ components. Instead, a patch is first compared to the two
first nodes in level 1 of the tree, then the branch providing the smallest
cost is followed and the operation is repeated at higher levels until
a leaf has been reached.
Using this balanced search tree reduces the complexity
of step \stepref{eq:gauss_select}
to $\Oo(N P \bar{r} \log K)$.
\subsection{Speed-up via the restriction to a random subset of patches}
\begin{figure}[!t]
  \centering%
  \subfigure[Regular patch subsampling at $(i_0, j_0)$]{\includegraphics[width=\linewidth, viewport=1 1 399 103, clip]{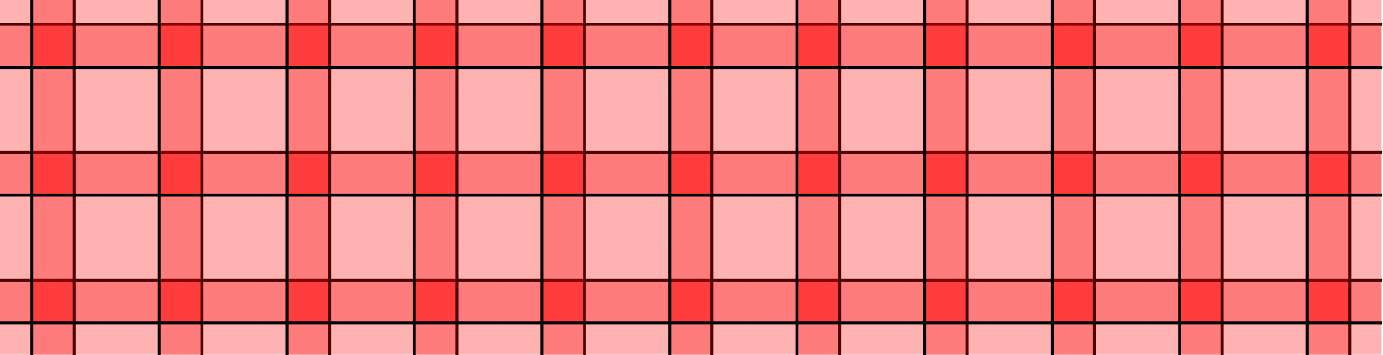}}\\
  \subfigure[Stochastic patch subsampling at $(i, j)$] {\includegraphics[width=\linewidth, viewport=1 1 399 103, clip]{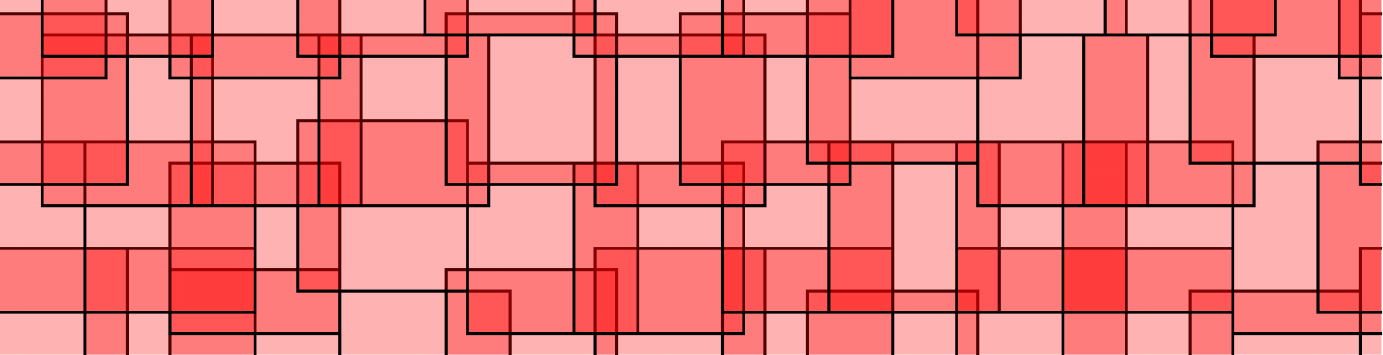}}\\
  \caption{
    Illustration of patch subsampling. Instead of extracting all patches,
    only a subset of patches is extracted either
    (a) regularly or (b) with
    some randomizations. Patches are represented by $8\!\times\!8$ squares and the
    red intensity represents the number of patches overlapping the corresponding pixel.
  }
  \label{fig:patchsubsampling}
\end{figure}
The simplest and most effective proposed acceleration consists of
subsampling the set $\Ii$ of $N$ patches to improve the complexity of
the four most time-consuming steps, see Table~\ref{tab:profiling}.
One approach, followed by BM3D \cite{Dabov07},
consists of restricting the set $\Ii$ to locations on a regular grid with spacing $s\in[1,\sqrt{P}]$
pixels in both directions, leading to a reduction of complexity by a factor $s^2$.
We refer to this approach as the regular patch subsampling.
A direct consequence is that $|\Ii|\!=\!N / s^2$ and the
complexity is divided by $s^2$.
However, we observed that this strategy consistently
creates
blocky artifacts revealing the regularity of the extraction pattern.
A random sampling approach, called "jittering", used in the computer graphics community \cite{cook1986stochastic} is preferable to limit this effect. This procedure
ensures that each pixel is covered by at least one patch. The location $(i_0, j_0)$ of a point of the grid undergoes a random perturbation,
giving a new location $(i,j)$ such that
\begin{align}
  &
  i_0 - \left\lfloor \tfrac{\sqrt{P}-s}{2} \right\rfloor
  \leq i \leq
  i_0 + \left\lfloor \tfrac{\sqrt{P}-s}{2} \right\rfloor
  \nonumber
  \\
  \qandq&
  j_0 - \left\lfloor \tfrac{\sqrt{P}-s}{2} \right\rfloor
  \leq j \leq
  j_0 + \left\lfloor \tfrac{\sqrt{P}-s}{2} \right\rfloor
  ,
\end{align}
where $\lfloor\cdot\rfloor$ denotes the flooring operation.
We found experimentally that independent and uniform perturbations
offered the best performance against all other tested strategies.
In addition, we also resample these positions at each
of the $T$ iterations
and add a (random) global
shift to ensure that all pixels have the same expected
number of patches covering them.

Figure \ref{fig:patchsubsampling} illustrates the difference between a regular grid
and a jittered grid of period $s\!=\!6$ for patches of size $P\!=\!8\!\times\!8$.
In both cases, all pixels are
covered by at least one patch, but the stochastic version
reveals an irregular pattern.

Nevertheless, when using random subsampling,
a major bottleneck occurs when $\Aa^t\Aa$ is not diagonal
because the inversion involved in eq.~\eqref{eq:qhs_optim_imest_sol} cannot be simplified
as in Alg.~\ref{alg:epll}.
Using a conjugate gradient is a practical solution but will
negate the reduction of complexity gained by using subsampling.
To the best of our knowledge, this is the main reason why patch subsampling has not been utilized to speed up EPLL.
Here, we follow a different path.
We opt for approximating the solution of the original problem (involving
all patches) rather than evaluating the solution
of an approximate problem (involving random subsample of patches).
More precisely, we speed up Alg.~\ref{alg:epll}
by replacing the complete set of indices
by the random subset of patches.
In this case, step \stepref{eq:patch_reprojection}
consists of averaging only this subset of overlapping restored patches.
This novel and nuanced idea avoids additional
overhead and attains dramatic complexity improvements compared to the standard
approach.

Experiments conducted on our validation dataset show that this strategy used with $s\!=\!6$
leads to an acceleration of about $36\times$ with less than a $0.2$dB drop in PSNR.
In comparison, for a similar drop of PSNR, the regular patch subsampling
can only achieve a $9\times$ acceleration with $s\!=\!3$.
\begin{figure}[!t]
  \centering%
  \subfigure{\includegraphics[width=\linewidth]{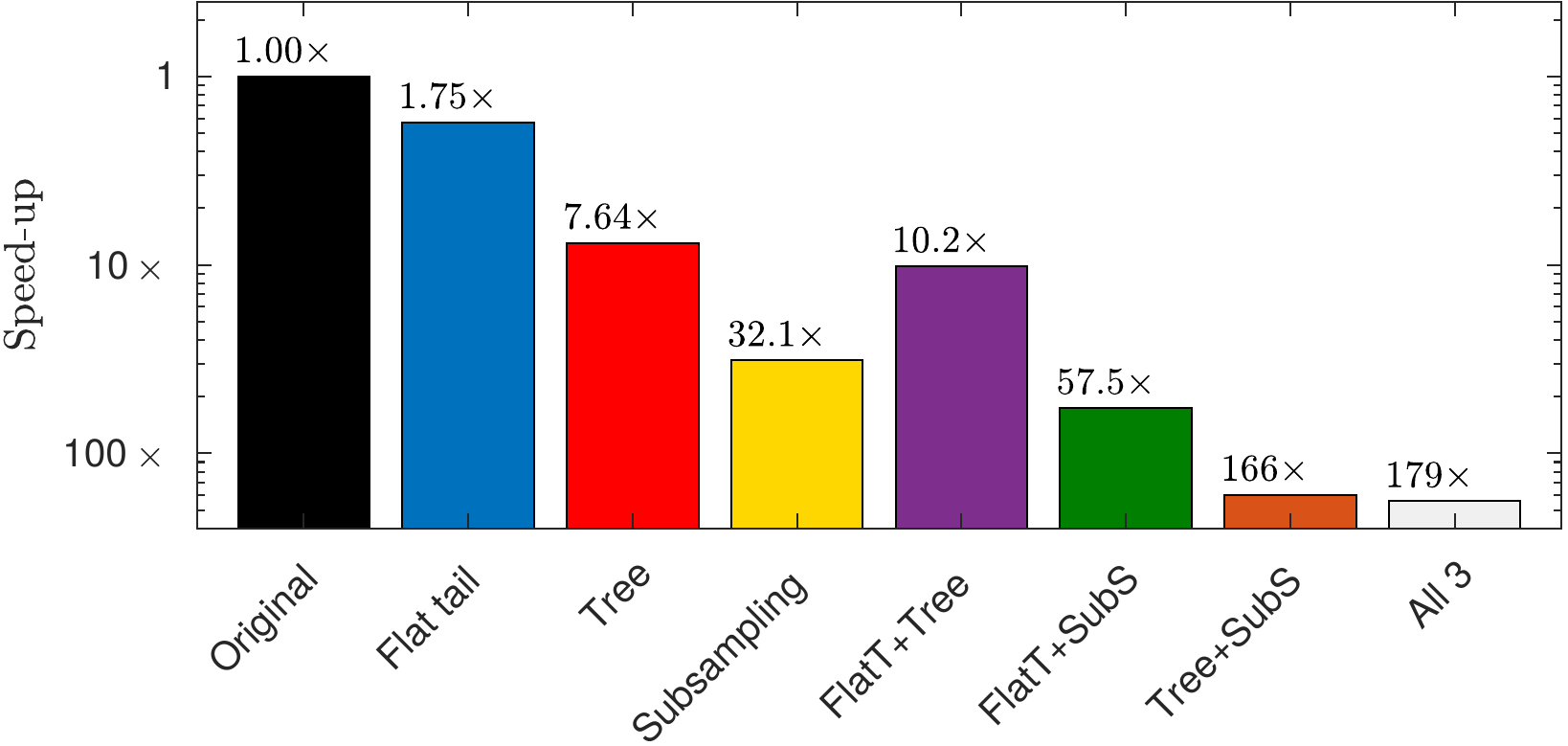}}\\
  \subfigure{\includegraphics[width=\linewidth]{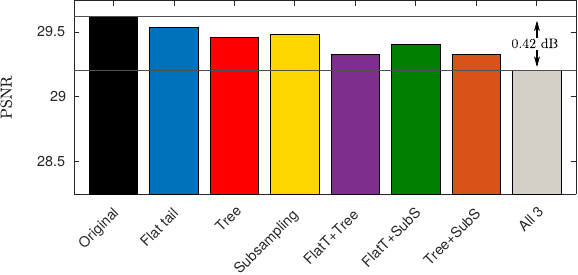}}\\
  \caption{%
    (left) Average speed-up and (right) average PSNR for our three accelerations,
    and all possible combinations of them,
    on the $40$ images of the BSDS validation set.
  }
  \label{fig:time_and_psnr}
\end{figure}
\subsection{Performance analysis}
Figure \ref{fig:time_and_psnr} shows the image restoration performance and speed-up obtained when
the three ingredients are applied separately or in combination. The results are averaged over
40 images from the test set of BSDS dataset \cite{martin2001database} that is set aside for
validation purposes. The speed-up is calculated with respect to the EPLLc implementation which
is labeled ``original'' in Fig.~\ref{fig:time_and_psnr}. Among the three ingredients, random subsampling or jittering (labeled ``subsampling'') leads to the largest speed-up ($32\times$), while the usage of the search
tree provides more than $7\times$ faster processing.
The average speed-up obtained when combining all three ingredients is around $179\times$ on our validation set
consisting of images of size $481\!\times\!321$,
for an average drop of PSNR less than $0.5$dB.
\section{Related methods}\label{sec:related}

To the best of our knowledge, there are only two other approaches
\cite{wang2014discriminative,rosenbaum2015return} that
have attempted to accelerate EPLL.
Unlike our approach, these methods focus on accelerating only one of the
steps of EPLL namely the \stepref{eq:gauss_select} step.
Both use machine learning techniques to
reduce its runtime.

In \cite{wang2014discriminative},
the authors use a binary decision tree
to approximate the mapping $\tilde \bz_i \mapsto k^*_i$
performed in step \stepref{eq:gauss_select}.
At each node $k$ and level $n$ of the tree, the patch $\tilde \bz_i$
is confronted with a linear separator in order to decide
if the recursion should continue on the left or right child given by
\begin{align}
  \dotp{\ba_k^n}{\tilde \bz_i} + b_k^n \geq 0
\end{align}
where $(\ba_k^n, b_k^n)$ are the parameters of the hyperplane
for the $k$-th node at level $n$.
These separators are trained offline
on all pairs of $(\tilde \bz_i, k^*_i)$ obtained after the first iteration of EPLL for a \textit{given} $\beta$ and
noise level $\sigma$.
Once a leaf has been reached, its index provides a first
estimate for the index $k^*_i$.
To reduce errors due to large variations among the neighboring pixels, this method further employs a Markov random fields on the resulting map of Gaussian components which runs in $\Oo(N K)$ complexity.
Hence, their overall approach reduces the complexity of step \stepref{eq:gauss_select} from $\Oo(N K P^2)$
to $\Oo(N (P D + K))$, where $D=12$ is the depth of the learned decision tree.

\begin{table*}[!t]
  \centering
  \caption{PSNR, SSIM and execution time on the BSDS test set (average on 60 images of size $481\!\times\!321$),
    and on six standard images (each of size $512\!\times\!512$) for the proposed FEPLL and FEPLL$'$,
    EPLL (with timing given for both EPLLm \cite{Zoran11} and our EPLLc),
    BM3D \cite{Dabov07}, CSF \cite{schmidt2014shrinkage} and RoG \cite{rosenbaum2015return}
    with 3 different levels of noise.}
  \begin{tabular}{ll ccccccc}
    \hline
    $\sigma$& Algo. & Berkeley & Barbara & Boat & Couple & Fingerprint & Lena & Mandrill\\
    \hline
    \hline
    \\[-1em]
    & & \multicolumn{7}{c}{PSNR/SSIM}\\
    \cline{3-9}\\[-0.9em]
    \multirow{7}{*}{$5$}
& FEPLL  			& 36.8 / .959  & 36.9 / .958  & 36.6 / .930  & 36.7 / .944  & 35.4 / .984  & 38.1 / .941  & 35.1 / .959 \\
& FEPLL$'$  		& 37.1 / .962  & 37.4 / .961  & 36.8 / .933  & 37.0 / .948  & 35.9 / .985  & 38.4 / .943  & 35.2 / .960 \\
&  EPLL 			& 37.3 / .963 & 37.6 / .962 & 36.8 / .933 & 37.3 / .949 & 36.5 / .987 & 38.6 / .944 & 35.2 / .960 \\
& RoG  					& 37.1 / .959  & 36.7 / .954  & 36.5 / .920  & 37.2 / .946  & 36.4 / .987  & 38.3 / .938  & 35.0 / .956 \\
&  BM3D  			& 37.3 / .962 & 38.3 / .964 & 37.3 / .939 & 37.4 / .949 & 36.5 / .987 & 38.7 / .944 & 35.2 / .959 \\
& CSF$_{3\times3}$   & 36.8 / .952  & 37.0 / .955  & 36.7 / .929  & 37.1 / .945  & 36.2 / .986  & 38.2 / .938  & 34.8 / .953 \\
\hline
\multirow{7}{*}{$20$}
& FEPLL  			& 29.1 / .812  & 29.1 / .853  & 30.1 / .802  & 29.9 / .812  & 27.7 / .909  & 32.3 / .863  & 26.4 / .786 \\
& FEPLL$'$  		& 29.3 / .831  & 29.7 / .869  & 30.4 / .815  & 30.2 / .827  & 28.1 / .920  & 32.4 / .865  & 26.6 / .805 \\
&  EPLL 			& 29.5 / .836  & 29.9 / .874  & 30.6 / .821  & 30.4 / .833  & 28.3 / .924  & 32.6 / .869  & 26.7 / .808 \\
& RoG  				& 29.3 / .828  & 28.4 / .838  & 30.4 / .815  & 30.2 / .826  & 28.3 / .922  & 32.5 / .866  & 26.5 / .796 \\
&  BM3D  			& 29.4 / .824  & 31.8 / .905  & 30.8 / .825  & 30.7 / .842  & 28.8 / .929  & 33.0 / .877  & 26.6 / .796 \\
& CSF$_{3\times3}$	& 29.0 / .805  & 28.3 / .822  & 30.2 / .802  & 29.9 / .812  & 28.0 / .917  & 31.8 / .838  & 26.1 / .779 \\
\hline
\multirow{7}{*}{$60$}
& FEPLL  				& 24.5 / .614  & 23.5 / .634  & 25.5 / .644  & 25.0 / .626  & 22.1 / .721  & 27.4 / .742  & 21.4 / .468 \\
& FEPLL$'$  			& 24.5 / .620  & 23.8 / .646  & 25.6 / .646  & 25.1 / .634  & 22.4 / .746  & 27.2 / .729  & 21.6 / .501 \\
&  EPLL 				& 24.8 / .631  & 24.0 / .658  & 25.8 / .658  & 25.3 / .646  & 22.6 / .754  & 27.6 / .746  & 21.7 / .506 \\
& RoG  					& 24.6 / .622  & 23.3 / .624  & 25.6 / .653  & 25.1 / .638  & 22.4 / .739  & 27.3 / .744  & 21.5 / .480 \\
&  BM3D  				& 24.8 / .637  & 26.3 / .756  & 25.9 / .671  & 25.5 / .662  & 23.7 / .800  & 28.2 / .775  & 21.7 / .501 \\
&  CSF$_{3\times3}$ 	& 22.0 / .489  & 21.4 / .487  & 22.7 / .504  & 22.5 / .503  & 21.2 / .727  & 23.4 / .509  & 20.3 / .467 \\
\hline
\\[-1em]
& & \multicolumn{7}{c}{Time (in seconds)}\\
\cline{3-9}\\[-0.9em]
& FEPLL  	& \phantom{00}0.27  & \phantom{00}0.42  & \phantom{00}0.40  & \phantom{00}0.41  & \phantom{00}0.38  & \phantom{00}0.43  & \phantom{00}0.42 \\
& FEPLL$'$  	& \phantom{00}0.96  & \phantom{00}1.91  & \phantom{00}1.90  & \phantom{00}1.90  & \phantom{00}1.90  & \phantom{00}1.90  & \phantom{00}1.90 \\
& EPLLc & \phantom{0}44.86 & \phantom{0}78.66  & \phantom{0}78.71  & \phantom{0}78.78  & \phantom{0}78.17  & \phantom{0}78.13  & \phantom{0}78.34  \\
& EPLLm  & \phantom{0}82.68 & 145.15  & 143.68  & 144.30  & 144.13  & 144.15  & 143.86  \\
& RoG  	 & \phantom{00}1.19  & \phantom{00}1.99  & \phantom{00}1.97  & \phantom{00}1.96  & \phantom{00}1.94  & \phantom{00}1.98  & \phantom{00}1.96 \\
& BM3D   & \phantom{00}1.64 & \phantom{00}2.79  & \phantom{00}2.91  & \phantom{00}2.83  & \phantom{00}2.45  & \phantom{00}2.87  & \phantom{00}2.84  \\
& CSF$_{3\times3}$ & \phantom{00}0.87  & \phantom{00}1.09  & \phantom{00}1.09  & \phantom{00}1.13  & \phantom{00}1.09  & \phantom{00}1.09  & \phantom{00}1.08 \\
& CSF$_{3\times3}^{\text{gpu}}$  & \phantom{00}0.41  & \phantom{00}0.41  & \phantom{00}0.41  & \phantom{00}0.41  & \phantom{00}0.41  & \phantom{00}0.40  & \phantom{00}0.41 \\
\hline
  \end{tabular}
  \label{tab:denoising}
  \vskip1em
\end{table*}

In \cite{rosenbaum2015return},
the authors approximate the \stepref{eq:gauss_select} step,
by using a gating (feed-forward) network with one hidden layer
\begin{align}\label{eq:rog}
  \tilde \bz_i \mapsto \iota^{k} \!+\! \sum_{j = 1}^Q \biggl(\log \nu^{k}_j
  \!+\!
  \omega^k_j (\tilde \bd_i)_j^2\biggr)
  \qwithq
  \tilde \bd_i = \bV^t \tilde \bz_i
\end{align}
where $Q$ is the size of the hidden layer.
 The matrix $\bV \in \RR^{P \times Q}$ encodes the weights of the first layer,
$\omega^k$ corresponds to the weights of the hidden layer
and they are learned discriminatively
to approximate the exact {\it posterior} probability:
\begin{align}\label{eq:epll_score}
  \tilde \bz_i \mapsto \iota^{k} \!+\! \sum_{j = 1}^P \biggl(\log \nu^{k}_j
  \!+\!
  \frac{(\tilde \bc_i^k)_j^2}{\nu^{k}_j}\biggr)
  \qwithq
  \tilde \bc_i^k = \bU_k^t \tilde \bz_i
\end{align}
that we encounter in eq.~\eqref{eq:projection_eigenspace} and \eqref{eq:selection_eigenspace}.
Theoretically, a new network will need to be trained
for each type of degradations, noise levels
and choices of $\beta$ (recall that $\nu^{k}_j = (\bS_k)_{jj} + \frac{1}{\beta}$).
However, the findings of
\cite{rosenbaum2015return} indicate that
applying a network learned on clean patches and with $\frac{1}{\beta} = 0$ is effective
regardless of the type of degradation or the value of $\beta$.
Their main advantage can be highlighted by comparing
eq.~\eqref{eq:rog} and \eqref{eq:epll_score} where complexity is reduced
from $\Oo(N K P^2)$ to $\Oo(N Q (K + P))$.
The authors utilize this benefit by choosing $Q = 100$.

Unlike these two approaches,
our method does not try to learn the Gaussian selection rule directly
(which depends on both the noise level through $1/\beta$ and the {\it prior} model through the GMM). Instead, we
simply define a hierarchical
organization of the covariance matrices $\bSigma_k$. In other words,
while the two other approaches try to infer the {\it posterior} probabilities
(or directly the maximum {\it a posteriori}),
our approach provides an approximation to the {\it prior} model.
During runtime,
this approximation of the {\it prior} is used in the {\it posterior}
for the Gaussian selection task.
Please note that the value of $\beta$ does not play a role in determining
the {\it prior}.
This allows us to use the same search tree
independently of the noise level, degradations, {\it etc.}
Given that the main advantage of EPLL is that the same model can be used
for any type of degradations, it is important that this
 property remains true for the accelerated version.
Last but not least, the training of our search tree takes
a few minutes while the training steps for the above mentioned approach
take from several hours to a few days \cite{wang2014discriminative}.

In the next section, we show that
our proposed accelerations produce restoration results with comparable quality
to competing methods while requiring a smaller amount of time.
\section{Numerical experiments}
\label{sec:numerical}
In this section, we present the results obtained
on various image restoration tasks. Our experiments were conducted on
standard images of size $512 \times 512$
such as Barbara, Boat, Couple,
Fingerprint, Lena, Mandrill and on
60 test images of size $481\!\times\!321$ from the  Berkeley Segmentation Dataset (BSDS) \cite{martin2001database}
(the original BSDS test set contains 100 images, the other 40 was
used for validation purposes while setting parameters $\rho$ and $s$).
For denoising, we compare the performance of our fast EPLL (FEPLL) to the original EPLL
algorithm \cite{Zoran11} and BM3D \cite{Dabov07}. For the original EPLL, we have
included timing results given by our own \texttt{MATLAB/C} implementation (EPLLc) and the
\texttt{MATLAB} implementation provided by the authors (EPLLm). We also compare our restoration
performance and runtime against other fast restoration methods introduced to
achieve competitive trade-off between runtime efficiency and image quality.
These methods include RoG \cite{rosenbaum2015return}
(a method accelerating EPLL based on feedforward networks described in Sec.~\ref{sec:related}),
and CSF \cite{schmidt2014shrinkage}
(a fast restoration technique using random field-based architecture).

For deblurring experiments, we additionally compare
with field-of-experts (FoE)-based non-blind deconvolution \cite{chen2014insights}
denoted as iPiano. We contacted the corresponding author of \cite{wang2014discriminative}
and got confirmation that the implementation of their algorithm
(briefly described in \ref{sec:related}) is not publicly available.
Due to certain missing technical details, we were
unable to reimplement it faithfully. However, the results reported in \cite{wang2014discriminative}
indicate that their algorithm performs in par with BM3D in terms of both PSNR and time.
Hence, BM3D results can be used as a faithful proxy for the expected performance of Wang et al.'s
algorithm \cite{wang2014discriminative}.

To explicitly illustrate the quality vs. runtime tradeoff of FEPLL,
we include results obtained using a slightly slower version of FEPLL referred to as FEPLL$'$,
that does not
use the balanced search tree and uses a flat tail spectrum approximation with $\rho = 0.98$.
Please note that FEPLL$'$ is not meant to be better or worse than FEPLL,
it is just another version running at a different PSNR/time tradeoff
which allows us to compare our algorithm to others operating
in different playing fields.

Finally, to illustrate the versatility of FEPLL,
we also include results for other inverse problems such
as devignetting, super-resolution,
and inpainting.
\begin{figure*}[!t]
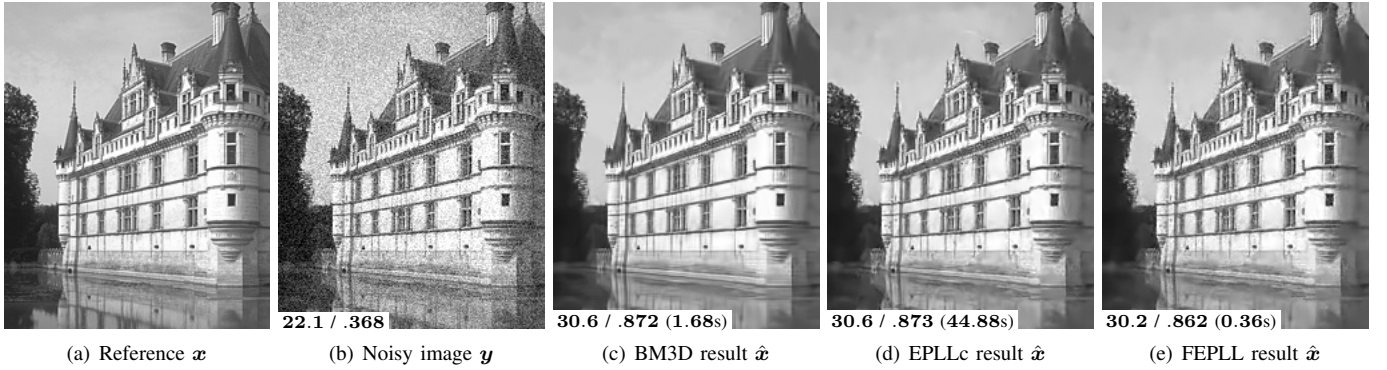

  \centering%
  \subfigure[Reference $\bx$]{%
    \includegraphicsnopsnr{0.195\linewidth}{viewport=0 170 200 415,clip}{denoising/subimg1}%
  }\hfill%
  \subfigure[Noisy image $\by$]{%
    \includegraphicspsnr{0.195\linewidth}{viewport=0 170 200 415,clip}{denoising/subimg2}%
                        {22.1}{.368}%

  }\hfill%
  \subfigure[BM3D result $\hat \bx$]{%
    \includegraphicspsnrtime{0.195\linewidth}{viewport=0 170 200 415,clip}{denoising/subimg5}%
                        {30.6}{.872}{1.68}%

  }\hfill%
  \subfigure[EPLLc result $\hat \bx$]{%
    \includegraphicspsnrtime{0.195\linewidth}{viewport=0 170 200 415,clip}{denoising/subimg4}%
                        {30.6}{.873}{44.88}%

  }\hfill%
  \subfigure[FEPLL result $\hat \bx$]{%
    \includegraphicspsnrtime{0.195\linewidth}{viewport=0 170 200 415,clip}{denoising/subimg3}%
                        {30.2}{.862}{0.36}%
  }%
  \\
  \caption{Illustration of a denoising problem
    with noise standard deviation $\sigma = 20$.
    Part of: (a) the original image
    (b) its noisy version
    (c-e) denoised results of competitive methods with PSNR/SSIM and time (inset).
  }
  \label{fig:denoising}
  \vskip1em
\end{figure*}
\paragraph*{Parameter settings}
\label{para:param_setting}
In our experiments, we use patches of size $P\!=\!8\!\times\!8$,
and the GMM provided by Zoran {\it et al.}~\cite{Zoran11}
with $K\!=\!200$ components.
The 200-components GMM
is progressively collapsed into smaller GMMs with $K\!=\!64, 32, 16, 8, 4, 2$ and $1$,
and then all Gaussians of the tree are modified offline by flat-tail approximations with $\rho\!=\!0.95$. The
final estimate for the restored image is obtained after 5 iterations of our
algorithm with $\beta$ set to ${\lambda}{\sigma^{-2}} \{ 1, 4, 8, 16, 32 \}$
where $\lambda\!=\!\min\{ N^{-1} {\norm{\Aa^t \Aa}_F^2}{\norm{\Aa}^{-2}_2}, 250 \sigma^2 \}$.
For denoising, where $\Aa$ is identity, $\lambda\!=\!1$ which boils down to the setting used by Zoran {\it et al.}~\cite{Zoran11}.
For inverse problems, we found that the initialization
$\hat \bx = (\Aa^t \Aa + 0.2 \sigma^2 / \lambda \nabla)^{-1} \Aa^t \by$,
with $\nabla$ the image Laplacian,
provides relevant solutions whatever
the linear operator $\Aa$ and the noise level $\sigma^2$.
While the authors of \cite{Zoran11} do not provide any further direction for
setting $\beta$ and the initialization in general inverse problems,
our proposed setting leads to competitive solutions irrespective
of $\Aa$ and $\sigma^2$.
For BM3D \cite{Dabov07}, EPLLm \cite{Zoran11}, RoG \cite{rosenbaum2015return},
CSF \cite{schmidt2014shrinkage} and iPiano \cite{chen2014insights}
we use the implementations provided by the original authors and use
the default parameters prescribed by them.
\begin{figure}
  \subfigure[]{\includegraphics[width=\linewidth,viewport=0 33 370 199,clip]{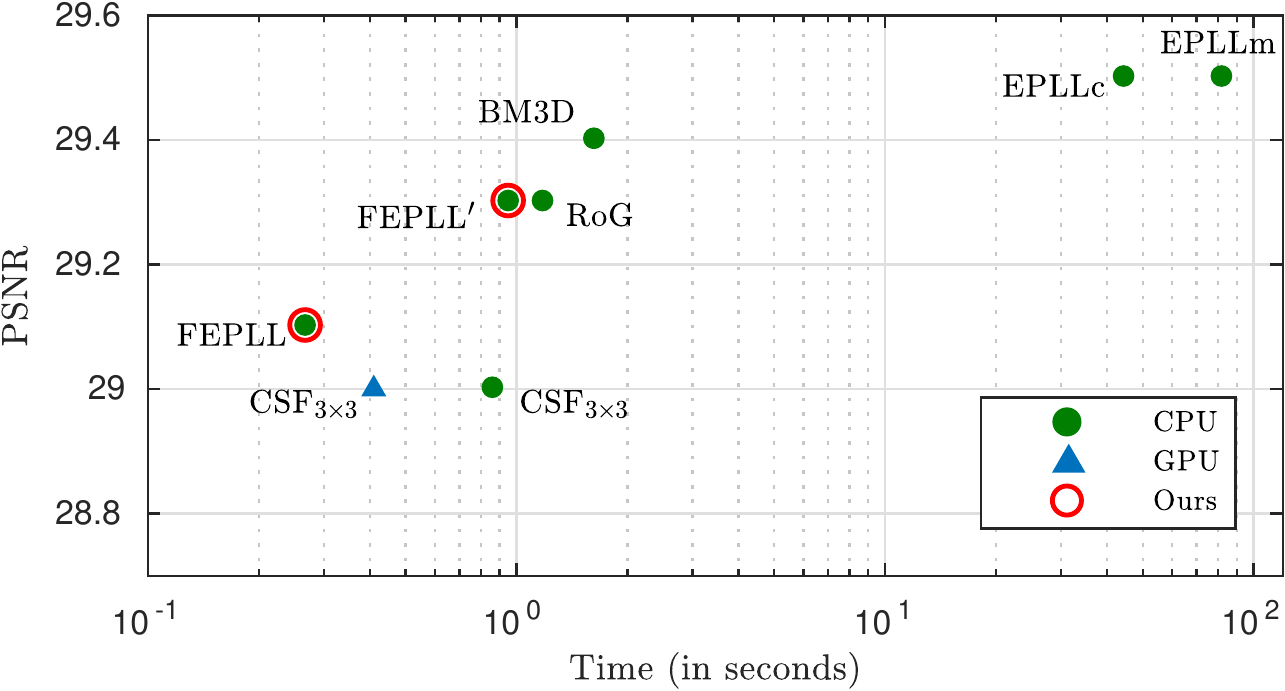}}
  \subfigure[]{\includegraphics[width=\linewidth]{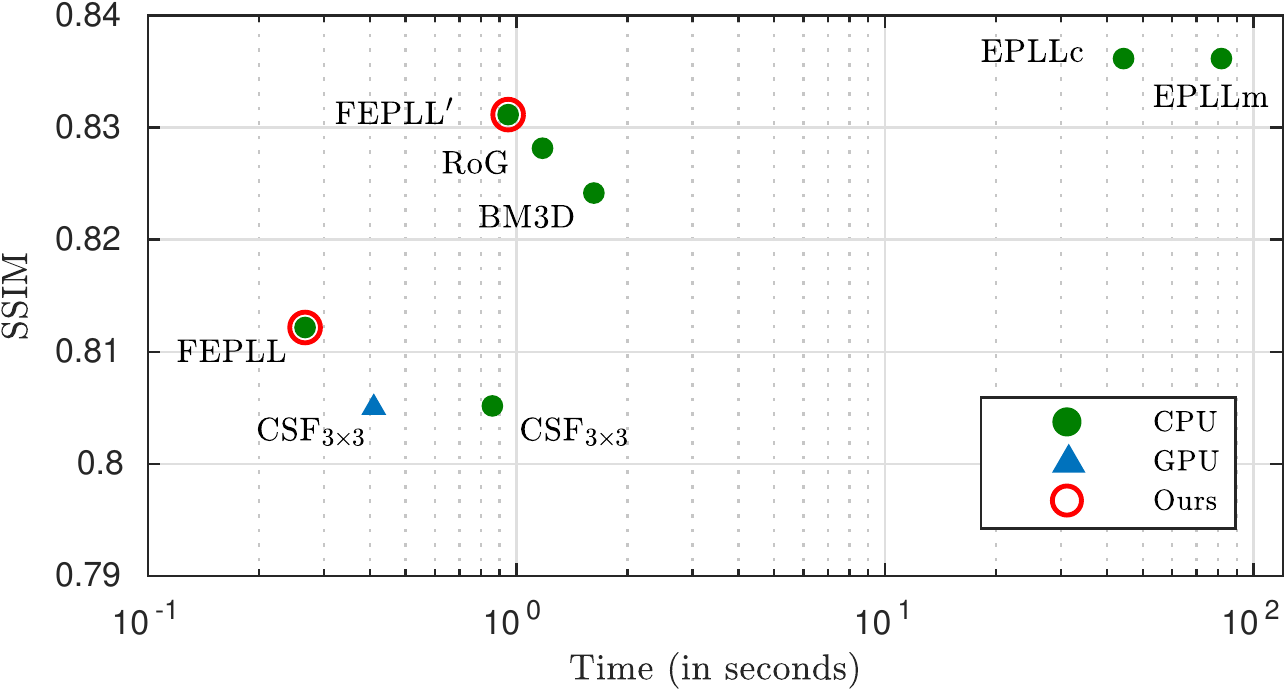}}
  \caption{(a) PSNR and (b) SSIM versus time for different restoration methods in a denoising
    problem with noise standard deviation $\sigma=20$.
    PSNR, SSIM and time are averaged on the 60 BSDS test images,
    each of size $481 \times 321$.
    Optimal methods tend to be in the top-left corner.
  }
  \label{fig:time_vs_psnr}
  \vskip1em
\end{figure}
\paragraph*{Denoising}
Table \ref{tab:denoising} shows the quantitative performances of FEPLL on the denoising task
compared to EPLLm \cite{Zoran11}, EPLLc (our own \texttt{MATLAB/C} implementation),
RoG \cite{rosenbaum2015return}
BM3D \cite{Dabov07}
and CSF \cite{schmidt2014shrinkage}.
We evaluate the algorithms under low-, mid- and high-noise settings by using Gaussian noise of variance $5^2$, $20^2$ and $60^2$, respectively.
The result labeled ``Berkeley'' is an average over 60 images from the BSDS testing set \cite{martin2001database}.
Figures \ref{fig:time_vs_psnr} provide graphical representations of these performances in terms of PSNR/SSIM
versus computation time for the BSDS images for the noise variance setting $\sigma^2 = 20^2$.
On average, FEPLL results are 0.5dB below regular EPLL and BM3D; however, FEPLL is approximately 7 times faster than BM3D, 170-200 times faster than EPLLc and over 350 times faster than EPLLm. FEPLL outperforms the faster CSF algorithm in terms of both PSNR and time. In this case, FEPLL is even faster than the GPU accelerated version of CSF (CSF$_{\text{gpu}}$).
Our approach is 4 times faster than RoG
with a PSNR drop of 0.1-0.3dB.
Nevertheless, if we slow down FEPLL to FEPLL$'$,
we can easily neutralize this quality deficit while still being faster than RoG.
Note that these accelerations are obtained purely based on the approximations and \textit{no} parallel processing is used. Also, in most cases, a loss of 0.5dB may not affect the visual quality of the image. To illustrate this, we show a sample image denoised by BM3D, EPLL and FEPLL in
Fig.~\ref{fig:denoising}.
\begin{figure*}[!t]
  \centering%
  \subfigure[Reference $\bx$ / Blur kernel]{%
    \includegraphicsnopsnr{0.195\linewidth}{viewport=140 20 365 295,clip}{deblur_results/subimg1}%
    \setlength\fboxsep{0pt}%
    \setlength\fboxrule{2pt}%
    \hspace*{-0.08\linewidth}%
    \hspace*{-2pt}%
    \fcolorbox{white}{white}{\includegraphics[width=0.08\linewidth]{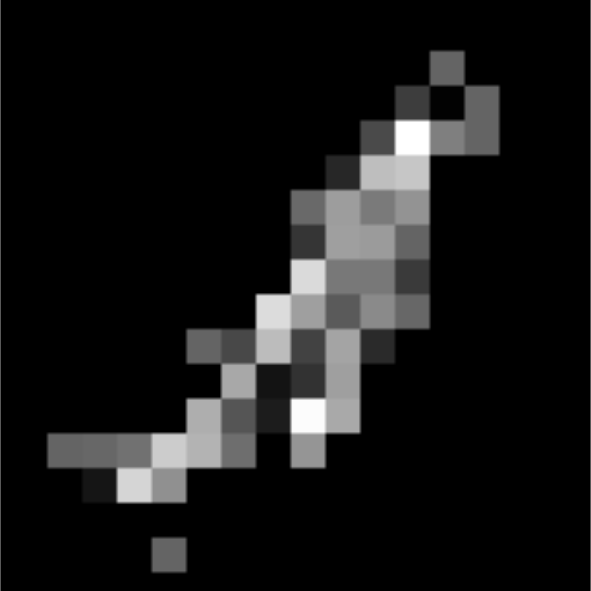}}%
    \hspace{-2pt}%
  }\hfill%
  \subfigure[Blurry image $\by$]{%
    \includegraphicspsnr{0.195\linewidth}{viewport=140 20 365 295,clip}{deblur_results/subimg2}%
                        {24.9}{.624}%
  }\hfill%
  \subfigure[CSF$_{\mathrm{pw}}$ result $\hat \bx$]{%
    \includegraphicspsnrtime{0.195\linewidth}{viewport=140 20 365 295,clip}{deblur_results/subimg4}%
                        {31.4}{.891}{0.17*}%

  }\hfill%
  \subfigure[RoG result $\hat \bx$]{%
    \includegraphicspsnrtime{0.195\linewidth}{viewport=140 20 365 295,clip}{deblur_results/subimg5}%
                        {32.2}{.910}{1.17}%

  }\hfill%
  \subfigure[FEPLL result $\hat \bx$]{%
    \includegraphicspsnrtime{0.195\linewidth}{viewport=140 20 365 295,clip}{deblur_results/subimg6}%
                        {32.7}{.924}{0.46}%
  }%
  \\
  \caption{Illustration of a deblurring problem
    with noise standard deviation $\sigma = 0.5$.
    Part of: (a) the original image
    and the blur kernel (inset),
    (b) blurry version
    (c-e) deblurred results of competitive methods with PSNR/SSIM and time (inset). The '*'
    indicates runtime on GPU while others are CPU times.
  }
  \label{fig:deblurring}
  \vskip1em
\end{figure*}
\paragraph*{Deblurring}
Table \ref{tab:deblurring} shows the performance of FEPLL when used for deblurring
as compared to
RoG \cite{rosenbaum2015return}, iPiano \cite{chen2014insights} and CSF  \cite{schmidt2014shrinkage}. For these
experiments, we use the blur kernel provided by Chen et al. \cite{chen2014insights} along
with their algorithm implementation. The results under the label ``Berkeley'' are averaged over 60 images
from the BSDS test dataset \cite{martin2001database}. The results labeled ``Classic'' is averaged
over the six standard images (Barbara, Boat, Couple,
Fingerprint, Lena and Mandrill). FEPLL consistently
outperforms its efficient competitors both in terms of quality and runtime.
Although the GPU version of CSF is faster, the restoration quality obtained
by CSF is 2-3dB lower than FEPLL. The proposed algorithm outperforms RoG
by 1-1.8dB while running 3 and 5 times faster on ``Berkeley'' and ``Classic'' datasets, respectively.

The superior qualitative performance of FEPLL is demonstrated in Fig. \ref{fig:deblurring}.
For brevity, we only include the deblurring results obtained from the top
competitors of FEPLL algorithm in terms of both quality and runtime. As
observed, FEPLL provides the best quality vs. runtime efficiency trade-off.
In contrast,
a deblurring procedure using the regular EPLL is around 350 times slower than FEPLL
with the original implementation \cite{Zoran11}. Specifically, on the sample image shown
in Fig. \ref{fig:deblurring}, EPLL provides a qualitatively similar result
(not shown in the figure) with a PSNR of 32.7 dB and SSIM of 0.922 in 142 seconds.

\begin{figure*}[!t]
  \centering%
  \includegraphicspsnr{0.325\linewidth}{viewport=2 2 479 319,clip}{train/subimg5}{11.1}{.662}\hfill%
  {%
    \includegraphicspsnr{0.325\linewidth}{viewport=2 2 479 319,clip}{train/subimg13}{20.8}{.598}%
    \setlength\fboxsep{0pt}%
    \setlength\fboxrule{2pt}%
    \hspace{-0.108\linewidth}%
    \hspace{-2pt}%
    \fcolorbox{white}{white}{\includegraphics[width=0.108\linewidth,viewport=2 2 479 319,clip]{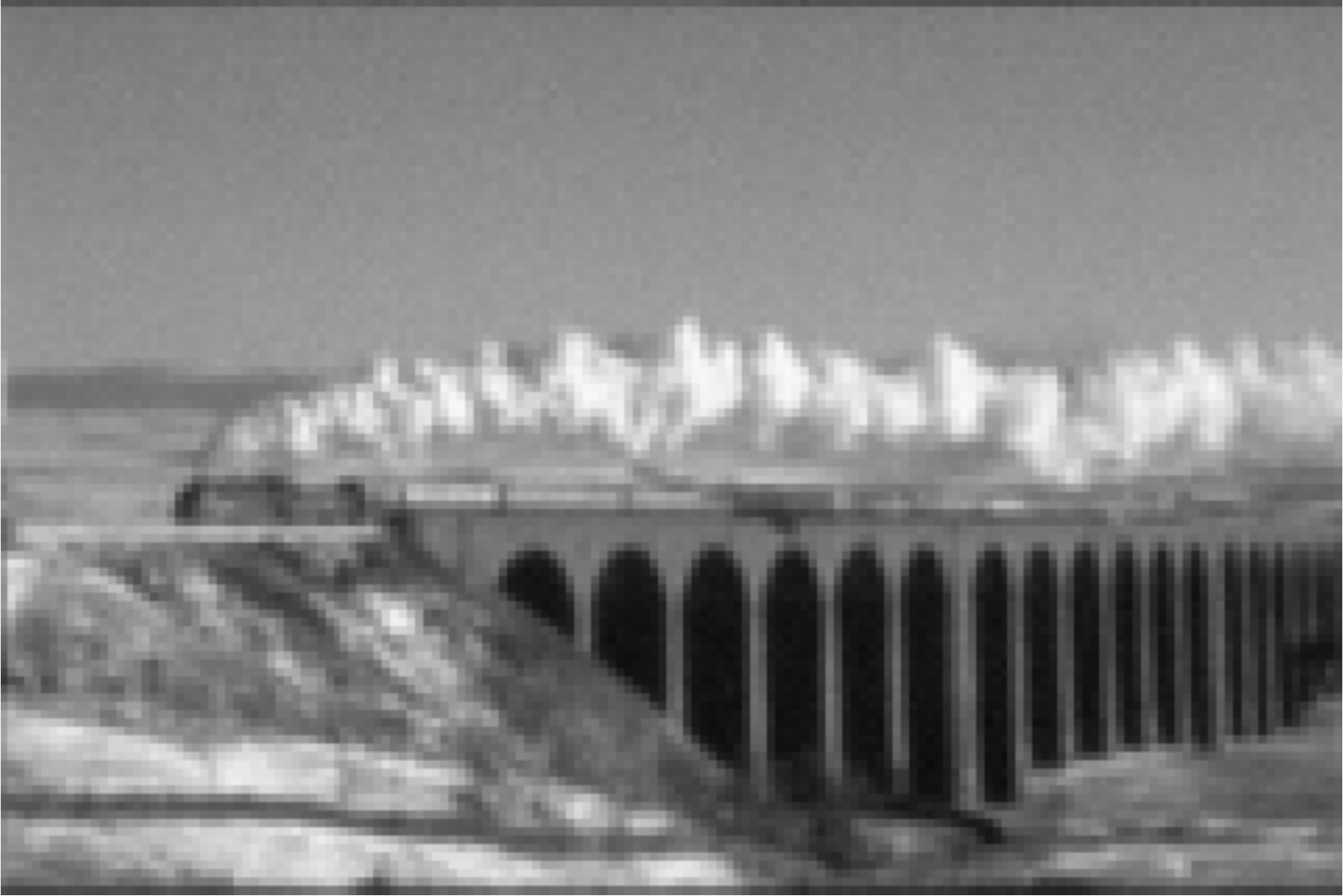}}%
    \hspace{-2pt}%
    }\hfill%
  \includegraphicspsnr{0.325\linewidth}{viewport=2 2 479 319,clip}{train/subimg17}{8.31}{.112}%
  \\[-0.5em]
  \subfigure[devignetting]{\includegraphicspsnrtime{0.325\linewidth}{viewport=2 2 479 319,clip}{train/subimg8}{36.8}{.972}{0.38}}\hfill%
  \subfigure[$\times 3$ super-resolution]{\includegraphicspsnrtime{0.325\linewidth}{viewport=2 2 479 319,clip}{train/subimg16}{23.3}{.738}{0.29}}\hfill%
  \subfigure[$50$\% inpainting]{%
    \includegraphicspsnrtime{0.325\linewidth}{viewport=2 2 479 319,clip}{train/subimg20}{27.0}{.905}{0.36}%
  }\\
  \caption{FEPLL on various inverse problems.
    All inputs contain Gaussian noise with $\sigma = 2$.
    Top row: (a) the observation in a devignetting problem, (b) the bi-cubic interpolation and the actual low-resolution size image (inset) in a $\times 3$ super-resolution problem and (c) the observation in an inpainting problem with $50\%$ of missing pixels shown in red.
    Bottom row: respective FEPLL results all obtained in less than $0.4$s.
  }
  \label{fig:inverse_problems}
  \vskip1em
\end{figure*}

\paragraph*{Other inverse problems} Unlike BM3D, EPLL and FEPLL are more versatile and handle
a wide range of inverse problems without any change in formulation.
In Fig.~\ref{fig:inverse_problems}, we show the results obtained by FEPLL on
problems such as (a) devignetting, which involves
a progressive loss of intensity,
(b) super-resolution and (c) inpainting.
To show the robustness of our method, the input
images of size $481\!\times\!321$ were degraded
with zero-mean Gaussian noise with $\sigma\!=\!2$.
All of the restoration results were obtained within/under 0.4 seconds and
with the same set of parameters explained above (cf. \textit{Parameter settings}).

\begin{table}[!t]
  \centering
  \caption{PSNR, SSIM and execution time on the BSDS test set (average of 60 images of size $481\!\times\!321$),
    and on standard images (average of 6 images of size $512\!\times\!512$) for the proposed FEPLL and FEPLL$'$,
    CSF \cite{schmidt2014shrinkage}, RoG \cite{rosenbaum2015return} and Chen et al.'s \cite{chen2014insights} method
    that is called \textit{iPiano} in their implementation. The blur kernel used is the one provided in along with
    iPiano implementation and noise is set to $\sigma = 0.5$.
  }
  \begin{tabular}{lcc@{}c@{}cc}
    \hline
    Algo.  & \multicolumn{2}{c}{Berkeley} && \multicolumn{2}{c}{Classic}\\
    \hline
    \hline
    & PSNR/SSIM & Time (s) & \hspace{1em} & PSNR/SSIM & Time (s)\\
    \cline{2-3}\cline{5-6}\\[-0.9em]
    iPiano	 &	29.5 / .824	&	29.53 && 29.9 / .848 & 59.10\\
    CSF$_{\text{pw}}$   &	30.2 / .875	&	0.50 (0.14*) && 30.5 / 0.870 & 0.47 (0.14*)\\
    RoG	&	31.3 / .897	&	1.19 && 31.8 / .915 & 2.07\\
    FEPLL 	&       33.1 / .928	&	0.40 && 32.8 / .931 & 0.46\\
    FEPLL' 	&	33.2 / .930	&	1.01 && 33.0 / .933 & 1.82\\
    \hline
  \end{tabular}
  \label{tab:deblurring}
  \vskip1em
\end{table}

\section{Conclusion}
\label{conclusion}
In this paper, we accelerate EPLL
by a factor greater than 100 with negligible loss of
image quality (less than $0.5$dB).
This is achieved by combining three independent
strategies: a flat tail approximation,
matching via a balanced search tree, and stochastic patch sampling.
We show that the proposed accelerations are effective
in denoising and deblurring problems, as well as in other inverse problems such as
super-resolution and devignetting.
An important distinction of the proposed accelerations is their genericity: the accelerated EPLL prior can be applied to many restoration tasks and various signal-to-noise ratios, in contrast to existing accelerations based on learning techniques applied to specific conditions (such as image size, noise level, blur kernel, etc.) and that require an expensive re-training to address a different problem.

Since the speed-up is achieved solely by reducing the algorithmic complexity,
we believe that further inclusion of accelerations based on
parallelization and/or GPU implementations will allow for real-time video processing.
Moreover, the acceleration techniques introduced in this work are general strategies
that can be used to speed up other image restoration and/or related machine learning algorithms.
For reproducibility purposes, the code of FEPLL
is made available on GitHub\footnote{https://goo.gl/xjqKUA}.

\bibliographystyle{abbrv}
\balance

\end{document}